\documentclass[letterpaper]{article} 
\usepackage{aaai24}  
\usepackage{times}  
\usepackage{helvet}  
\usepackage{courier}  
\usepackage[hyphens]{url}  
\usepackage{graphicx} 
\usepackage{natbib}  
\usepackage{caption} 
\usepackage{algorithm}
\usepackage{algorithmic}

\usepackage{epsfig}
\usepackage{graphics}
\usepackage{amsmath}
\usepackage{amssymb}
\usepackage{color, colortbl}
\usepackage{amsmath,amsfonts,mathtools,amssymb}
\usepackage{algorithm,algorithmic}
\usepackage{array}
\usepackage{color,soul}
\usepackage{xcolor}
\usepackage{bm}

\usepackage{multirow}
\usepackage{booktabs}
\usepackage{subfig}
\usepackage{mwe}

\DeclareMathOperator{\N}{\mathcal{N}}
\DeclareMathOperator{\I}{\mathbf{I}}
\DeclareMathOperator{\x}{\mathbf{x}}
\DeclareMathOperator{\xw}{\Tilde{\mathbf{x}}}

\usepackage{soul}

\definecolor{myblue}{RGB}{66,133,244}
\definecolor{mygreen}{RGB}{51,168,83}
\definecolor{myyellow}{RGB}{251,188,3}
\definecolor{myred}{RGB}{234,67,53}
\definecolor{mygrey}{RGB}{95,99,104}
\definecolor{mypup}{RGB}{153,0,204}

\usepackage{newfloat}
\usepackage{listings}
\DeclareCaptionStyle{ruled}{labelfont=normalfont,labelsep=colon,strut=off} 
\floatstyle{ruled}
\newfloat{listing}{tb}{lst}{}
\floatname{listing}{Listing}
\title{NightRain: Nighttime Video Deraining via Adaptive-Rain-Removal and Adaptive-Correction}

\author {
Beibei Lin\textsuperscript{\rm 1},
Yeying Jin\textsuperscript{\rm 1},
Wending Yan\textsuperscript{\rm 2},
Wei Ye\textsuperscript{\rm 2},
Yuan Yuan\textsuperscript{\rm 2},
Shunli Zhang\textsuperscript{\rm 3},
Robby Tan\textsuperscript{\rm 1}
}

\affiliations{
    \textsuperscript{\rm 1}National University of Singapore\\
    \textsuperscript{\rm 2}Huawei International Pte Ltd\\
    \textsuperscript{\rm 3}Beijing Jiaotong University\\

   \{beibei.lin, e0178303\}@u.nus.edu, \{yan.wending, yewei10, yuanyuan10\}@huawei.com, \\ slzhang@bjtu.edu.cn, robby.tan@nus.edu.sg
}

\usepackage{bibentry}

\begin{document}

\maketitle

\begin{abstract}
Existing deep-learning-based methods for nighttime video deraining rely on synthetic data due to the absence of real-world paired data. However, the intricacies of the real world, particularly with the presence of light effects and low-light regions affected by noise, create significant domain gaps, hampering synthetic-trained models in removing rain streaks properly and leading to over-saturation and color shifts. Motivated by this, we introduce NightRain, a novel nighttime video deraining method with adaptive-rain-removal and adaptive-correction. Our adaptive-rain-removal uses unlabeled rain videos to enable our model to derain real-world rain videos, particularly in regions affected by complex light effects. The idea is to allow our model to obtain rain-free regions based on the confidence scores. Once rain-free regions and the corresponding regions from our input are obtained, we can have region-based paired real data. These paired data are used to train our model using a teacher-student framework, allowing the model to iteratively learn from less challenging regions to more challenging regions. Our adaptive-correction aims to rectify errors in our model's predictions, such as over-saturation and color shifts. The idea is to learn from clear night input training videos based on the differences or distance between those input videos and their corresponding predictions. Our model learns from these differences, compelling our model to correct the errors. From extensive experiments, our method demonstrates state-of-the-art performance. It achieves a PSNR of 26.73dB, surpassing existing nighttime video deraining methods by a substantial margin of 13.7\%.
\end{abstract}

\section{Introduction}

\begin{figure}[]
{\includegraphics[width=2.7cm, height=2.7cm]{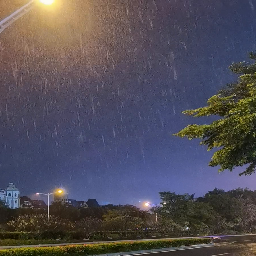}}\hspace{1pt}
{\includegraphics[width=2.7cm, height=2.7cm]{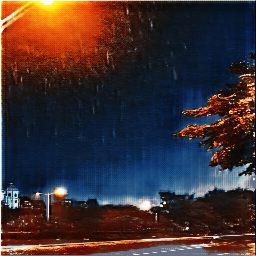}}\hspace{1pt}
{\includegraphics[width=2.7cm, height=2.7cm]{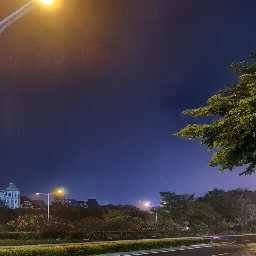}}\hspace{1pt}

\subfloat[Inputs]
{\includegraphics[width=2.7cm, height=2.7cm]{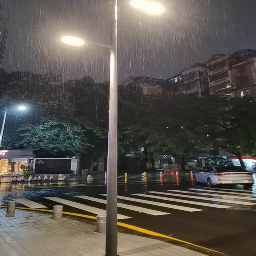}}\hspace{1pt}
\subfloat[MetaRain]
{\includegraphics[width=2.7cm, height=2.7cm]{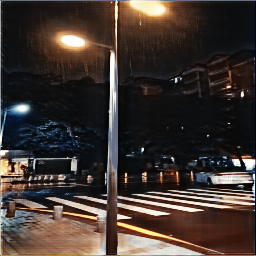}}\hspace{1pt}
\subfloat[\textbf{Ours}]
{\includegraphics[width=2.7cm, height=2.7cm]{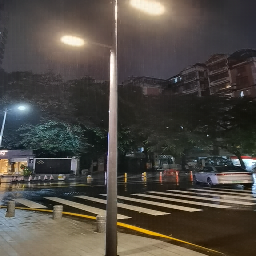}}\hspace{1pt}

\caption{Qualitative results on real-world nighttime rain videos. 
		First column: Input image. Second column: MetaRain's results~\cite{patil2022video}. Third column: Our results. Zoom-in for better visualization.
}
\label{fig_trailer}
\end{figure}

Nighttime rain videos are adversely affected by several factors, including low light, nighttime light effects (e.g., glow, glare, and uneven light distribution), rain streaks, etc.
Existing nighttime video deraining methods (i.e.,~\cite{patil2022video, patil2022dual}) rely on paired data for training. Due to the lack of real-world paired data, these methods are trained using synthetic datasets. 
However, complex nighttime light effects introduce a significant domain gap between synthetic and real-world rain streak distribution. 
Consequently, these methods struggle to effectively remove real-world rain streaks, particularly in the presence of complex light effects.
Moreover, models trained on synthetic datasets tend to produce over-saturated and color-shifted results. As shown in Fig.~\ref{fig_trailer}, existing nighttime video deraining methods yield over-saturated light regions and color-shifts in some regions.

Existing daytime video deraining methods \cite{zhang2022enhanced, kulkarni2021progressive, wang2022rethinking} can be retrained to remove nighttime rain streaks. However, these methods are designed for daytime scenes that do not have complex lighting conditions and the significant presence of noise in low-light regions. Thus, they struggle to handle rain streaks in nighttime scenes.
Existing single-image deraining methods \cite{shi2020joint, tang2020modified, chen2023learning, jiang2020multi} can be adapted to remove rain streaks by incorporating alignment and temporal smoothness priors. Nonetheless, they also encounter challenges from nighttime light effects and low-light regions. Additionally, alignment becomes problematic in the presence of intricate lighting conditions, rain streaks, and noise in low-light regions.

To address these challenges, we introduce \emph{NightRain}, a novel nighttime video deraining method with two main module: {\bf adaptive-rain-removal} and {\bf adaptive-correction}.
%
Note that, we exclude night videos with the rain veiling effect \cite{jin2023enhancing, jin2022unsupervised} (or rain accumulation) since these videos typically do not exhibit prominent rain streaks.

In our NightRain, our adaptive-rain-removal leverages unlabeled rain videos to enhance our model's deraining ability in real-world videos, particularly in low-light regions or  regions affected by complex light effects.
Specifically, we employ a pretrained model as a teacher model to generate rain-free predictions based on confidence scores. 
These high-confidence predictions primarily arise from less challenging regions, such as less pronounced low-light regions or regions with slight light effects.
In contrast, low-confidence predictions tend to result in inaccurate restorations because their corresponding inputs are more likely influenced by complex light effects or contain significantly low light regions.
High-confidence predictions and their corresponding inputs are then used as region-based paired real data to train a student model. Through iteratively transferring knowledge from the student model to the teacher model, our teacher model gains the ability to handle more challenging inputs, resulting in improved performance.
As shown in Fig. \ref{fig_trailer}, our model effectively removes rain streaks even in the presence of light effects.
Our adaptive-rain-removal operates as an unsupervised method as it does not require ground-truths.

Our adaptive-correction aims to rectify errors in our model's predictions, such as over-saturation and color shifts.
To achieve this, we utilize our video diffusion model to generate predictions from real-world nighttime clear videos. The discrepancies between clear input videos and our predictions essentially represent these errors. Thus, we take these predictions and their corresponding inputs as paired data to continually refine our model.
Through iterative error correction, our model is compelled to learn a more comprehensive nighttime clear distribution, leading to more precise predictions.
As shown in Fig. \ref{fig_trailer}, our method removes real-world rain streaks without introducing over-saturation and color-shifted problems.
Our adaptive-correction operates as an unsupervised method as we only use nighttime clear videos as inputs.

As a summary, our main contributions as follows:
\begin{enumerate}
	\item 
	We introduce adaptive-rain-removal to enable our model to derain real-world nighttime rain videos, especially in low-light regions or regions affected by complex light effects.
    By learning high-confidence predictions, our adaptive-rain-removal process forces our model to iteratively address more challenging regions, leading to effective real-world rain streak removal.  

	\item  
	We propose adaptive-correction to address over-saturated and color-shifted problems. By learning from differences between clear night videos and their corresponding predictions, our adaptive-correction continually rectifies errors in our models' predictions, resulting in more accurate restoration.

	\item 
    Experimental results demonstrate that our method not only achieves state-of-the-art restoration performance on the SynNightRain dataset, improving by 13.7\%, but also effectively removes real-world rain streaks even in the presence of light effects.

\end{enumerate}

\section{Related Work}

\noindent \textbf{Video Deraining} 
Existing nighttime video deraining methods~\cite{patil2022dual,patil2022video} remove rain streaks by networks trained on synthetic datasets. 
They synthesize rain streaks on clear nighttime videos, and then train CNNs on the synthetic datasets. 
However, their methods suffer from the domain gap problem since there exists a significant domain gap between synthetic and real-world rain streak dsi. 
Another problem of existing video deraining methods is that they tend to over-suppress the input videos, leading to color-shifted and over-saturated results.

Self-learning daytime video deraining methods \cite{yang2020self, yan2021self} 
do not require paired data to train their network and may be applied for nighttime video deraining.
Specifically, their methods are based on an assumption that adjacent frames may contain rain-free information and can be used to estimate a rain-free current frame.
However, these methods cannot remove dense rain streaks since adjacent frames in heavy rain do not contain enough rain-free information.
Some supervised daytime video deraining methods \cite{zhang2022enhanced,kulkarni2021progressive,wang2022rethinking,wang2022real,zhuang2022uconnet,xue2020sequential,wang2019survey,islam2021video, su2023complex,li2023stanet} can be directly used to address nighttime rain streaks. However, these methods also suffer from the domain gap problem since they rely on synthetic datasets for training.

\noindent \textbf{Diffusion Models} 
Existing diffusion models \cite{ozdenizci2023restoring,harvey2022flexible,peebles2022scalable, saharia2022image, ma2023deepcache, Fang_2023_CVPR} show a remarkable success in different generative tasks. 
Ozdenizci et al. \cite{ozdenizci2023restoring} propose a 2D UNet-based diffusion model to address multiple weather degradation, i.e., snow, raindrop and rain. 
Harvey et al. \cite{harvey2022flexible} present a 3D UNet-based diffusion model for video completions. 
These diffusion models can be re-trained on nighttime rain streak removal datasets and then used to remove real-world nighttime rain streaks. 
However, due to the lack of real-world paired data, the network training depends on synthetic datasets. As a result, their predictions also suffer from the domain gap problem, leading to inaccurate restoration.

\section{Proposed Method: NightRain}
\begin{figure}[t] 
	\begin{center}  
		\includegraphics[width=1.0\linewidth]{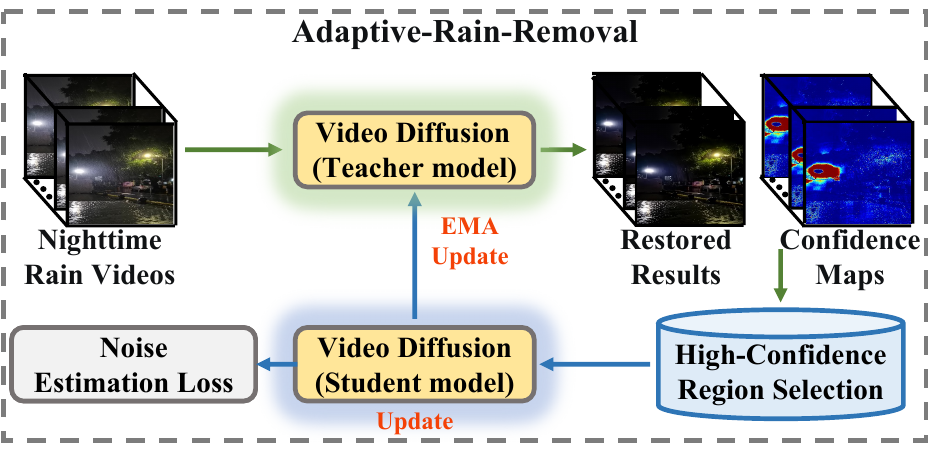} 
		\caption{
			Overview of our adaptive-rain-removal. 
			We pre-train a video diffusion model on synthetic nighttime deraining datasets as a teacher model. 
			Our adaptive-rain-removal utilizes the teacher model to generate predictions from real-world nighttime rain videos. We also generate confidence maps of rain streak removal within these predictions (green lines). 
			The high-confidence predictions with their corresponding inputs are then selected to train a student model, thus reducing the domain gap (blue lines). 
			Finally, we utilize Exponential Moving Average (EMA) to update our teacher model.
		}
		\label{Fig_overview_srr}
	\end{center}
\end{figure}

\begin{figure}[t] 
	\begin{center}  
		\includegraphics[width=1.0\linewidth]{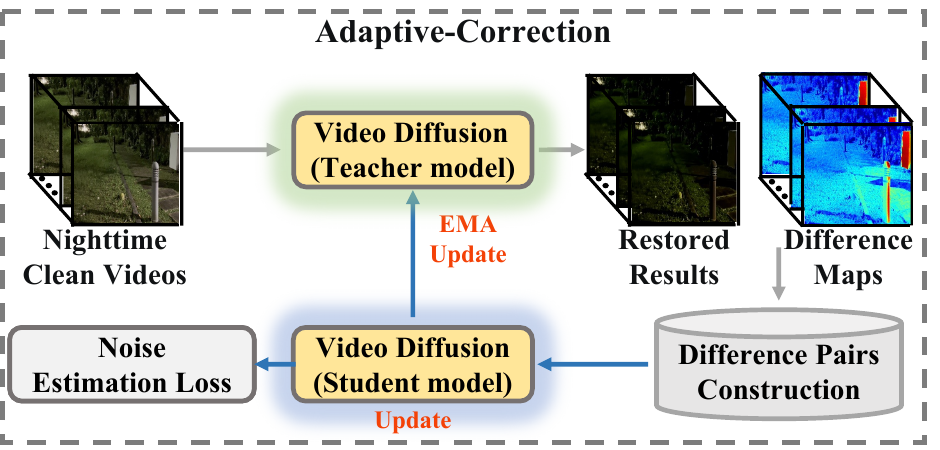} 
		\caption{
			Overview of our adaptive-correction. 
			We utilize our teacher model to generate predictions from nighttime clear videos (grey lines). The difference regions between clear videos and their corresponding predictions essentially represent errors produced by our model itself. We then use these difference pairs to train a student model, thus correcting our model's errors (blue lines). Finally, we utilize Exponential Moving Average (EMA) to update our teacher model.
		}
		\label{Fig_overview_sc}
	\end{center}
\end{figure}

Figs.~\ref{Fig_overview_srr} and~\ref{Fig_overview_sc} display the two main training processes for our NightRain: adaptive-rain-removal and adaptive-correction. The goal of our adaptive-rain-removal is to enable our model to remove rain streaks from real night videos without requiring ground-truths in its training phase.
The objective of our adaptive-correction is to enable our model to rectify errors in its predictions, such as over-saturation or color shifts, by exploiting nighttime clear videos. These training processes are separate from each other.

Initially, our adaptive-rain-removal and adaptive-correction rely on a pre-trained video diffusion model $w$ trained on a synthetic dataset $\mathbf{D}_{s} = \{ \mathbf{x}^{s}_{i}, \mathbf{y}^{s}_{i} \}^{N_{s}}_{i=1}$, where $\mathbf{x}^{s}_{i}$ and $\mathbf{y}^{s}_{i}$ are the $i$-th input and ground-truth videos, respectively. $N_{s}$ is the number of supervised videos. The pre-trained video diffusion model is implemented by a conditional video diffusion model. During the training phase, instead of sampling the input data $\x_0\sim q(\x_0)$ (standard diffusion model), we sample $(\x_0, \xw)\sim q(\x_0, \xw)$ to train our diffusion model, where $\x_0$ and $\xw$ denote clear and rain videos. The reverse process can be formulated as:
\begin{equation}
	\label{eqn_creverse1}
	p_{w}(\x_{0:T}|\xw) = p(\x_{T}) \prod_{t=1}^T p_{w}(\x_{t-1}\vert\x_t,\xw).
\end{equation}
Here, the inputs $\x$ and $\xw$ are concatenated channel-wise. During the inference phase, given a random Gaussian noise $\x_T\sim\N(\mathbf{0},\I)$ and a degraded video $\xw$, the clear video $\x_0$ can be generated by gradually reversing a diffusion process. $\x_{t-1}\sim p_{w}(\x_{t-1}\vert\x_t,\xw)$ is computed as:
\begin{equation}
	\label{eqn_creverse2}
	\begin{aligned}
		\x_{t-1} = &  \sqrt{\bar{\alpha}_{t-1}}\left(\frac{\x_t-\sqrt{1-\bar{\alpha}_t}\cdot\bm{\epsilon}_{w}(\x_t,\xw,t)}{\sqrt{\bar{\alpha}_t}}\right) \\
		&+ \sqrt{1-\bar{\alpha}_{t-1}}\cdot\bm{\epsilon}_{w}(\x_t,\xw,t).
	\end{aligned}
\end{equation}
where $\bm{\epsilon}_{w}(\cdot)$ and $\bm{\epsilon}_{w}(\x_t,\xw,t)$ are the noise estimation network and predicted noise vectors, respectively.

\begin{figure}
	\begin{center}  
		\includegraphics[width=1.0\linewidth]{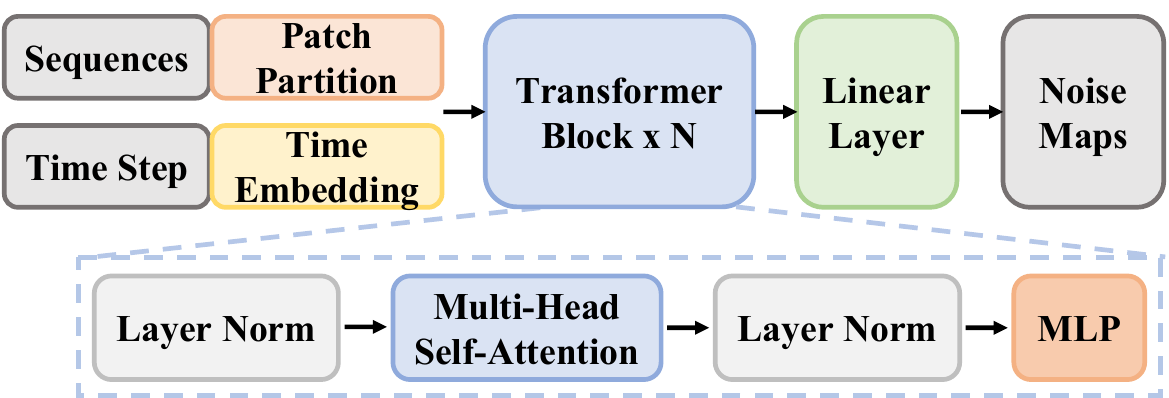} 
		\caption{
			Overview of our transformer-based noise estimation Network. 
			Given a sequence and a time step as inputs, we convert the sequence into patches and encode the time step to the time embedding.
			Subsequently, several transformer blocks are applied to extract global spatiotemporal cues.
			Finally, a linear layer is proposed to convert feature representations to noise maps.
		}
		\label{Fig_transformer}
	\end{center}
\end{figure}

In our method, we use a video transformer to implement our noise estimation network, which is shown in Fig.~\ref{Fig_transformer}. Given a sequence $X_{s}\in \mathbb{R}^{C_{s} \times T_{s} \times H_{s} \times W_{s}}$ and a time step $T_{s}\in \mathbb{R}^{1}$, where $C_{s}$ is the number of channels, $T_{s}$ is the length of sequences and ($H_{s}$,$W_{s}$) is the image size of each frame, we utilize a 3D convolution to partition the input sequences and utilize MLP to embed the time step. The two operations can be expressed as:
\begin{eqnarray}
	\label{Eqn_patch}
	X_{p} &=& F_{\rm ts\times ss\times ss}(X_{s}),\\
	\label{Eqn_time}
	T_{e} &=& {\rm MLP}(T_{s}),
\end{eqnarray}
where $X_{p} \in \mathbb{R}^{P \times C_{p}}$ and $T_{e}\in \mathbb{R}^{\times C_{p}}$ are patches and the time embedding. $F_{\rm ts\times ss\times ss}(\cdot)$ is a 3D convolution with the kernel size is $(\rm ts\times ss\times ss)$.

The patches $X_{p}$ and the time embedding $T_{e}$ are then injected into our transformer blocks, consisting of two LN layers, a Multi-head Self-Attention (MSA) module and an MLP layer, to extract global temporal and spatial information. 
We follow \cite{peebles2022scalable} to inject the time embedding for our diffusion model. 
Our transformer block first linearly transforms the time embedding to six parts, \emph{i.e.}, $s_{\rm msa}$, $g_{\rm msa}$, $c_{\rm msa}$,$s_{\rm mlp}$, $g_{\rm mlp}$ and $c_{\rm mlp}$. 
The size of each part is $ \mathbb{R}^{\times C_{p}}$. Then, a transformer block can be represented as:
\begin{small}
	\begin{eqnarray}
		Y^{\rm msa}_{p}&=& X_{p} + g_{\rm msa}*{\rm MSA}(\rm adaLN(X_{p}, s_{\rm msa}, c_{\rm msa})), \nonumber \\
		Y_{p}&=& Y^{\rm msa}_{p} + g_{\rm mlp}*{\rm MLP}(\rm adaLN(Y^{\rm msa}_{p}, s_{\rm mlp}, c_{\rm mlp})), \nonumber
	\end{eqnarray}
\end{small}
\hspace{-0.1cm}where $Y_{p}  \in \mathbb{R}^{P \times C_{p}}$ and $T_{e}\in \mathbb{R}^{\times C_{p}}$ is the output of a transformer block. $s_{\rm msa}$ and $s_{\rm mlp}$ are learnable shift parameters of adaptive layer norm (adaLN), while $c_{\rm msa}$ and $c_{\rm mlp}$ are learnable scale parameters of adaLN. $g_{\rm msa}$ and $g_{\rm mlp}$ are learnable scaling parameters of our transformer block. 
After using transformer blocks and a linear layer, we can obtain noise maps $\bm{\epsilon} \in \mathbb{R}^{\frac{C_{s}}{2} \times T_{s} \times H_{s} \times W_{s}}$.

Once the pretrained video diffusion model $w$ is obtained, our adaptive-rain-removal and adaptive-correction train this model further using real-world nighttime rain and clear videos.

\begin{figure}[t]
	\centering
	{\includegraphics[width=0.32\linewidth]{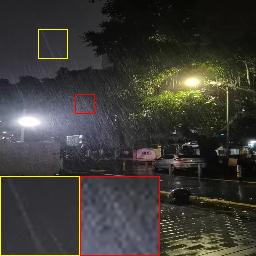}
		\includegraphics[width=0.32\linewidth]{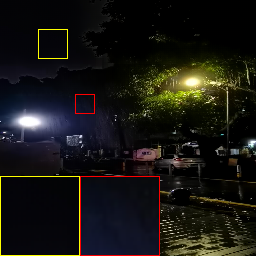}
		\includegraphics[width=0.32\linewidth]{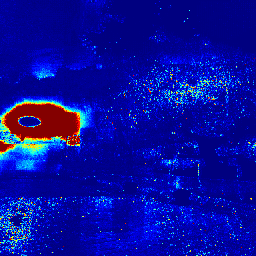}
	}
	{\includegraphics[width=0.32\linewidth]{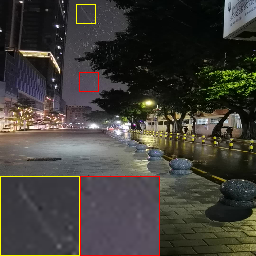}
		\includegraphics[width=0.32\linewidth]{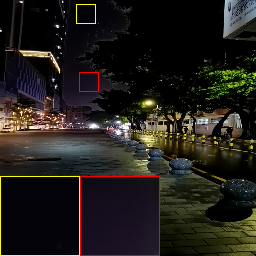}
		\includegraphics[width=0.32\linewidth]{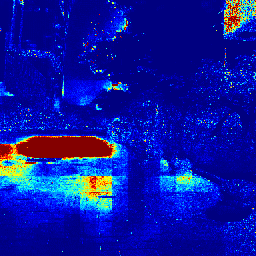}
	}\\
	\hspace{0.3cm} (a) Input  \hspace{0.8cm} (b) Predictions \hspace{0.5cm} (c) Confidence
	\\
	\caption{
		Visualization of predictions and confidence maps from our pretrained model. The red and blue regions are low-confidence and high-confidence predictions, respectively.
	}
	\label{fig_vis_self_enhance}
\end{figure}

\subsection{Adaptive-Rain-Removal}
\label{method_srr}
The core idea of adaptive-rain-removal is to improve our model's real-world deraining capability by learning from high-confidence predictions of actual rain videos. As shown in Fig. \ref{Fig_overview_srr}, our adaptive-rain-removal is implemented through a teacher-student framework. Hence, this adaptive-rain-removal process is unsupervised and thus does not require ground-truths.

We use the pretrained video diffusion model $w$ to initialize the teacher network $w_{T}$ and the student network $w_{S}$.
Given an unlabeled rain dataset $\mathbf{D}_{ur} = \{ \mathbf{x}^{ur}_{i}\}^{N_{\rm ur}}_{i=1}$, where $\mathbf{x}^{\rm ur}_{i}$ is the $i$-th rain video and $N_{\rm ur}$ is the number of rain videos, we utilize the teacher model $w_{T}$ to generate predictions $\mathbf{y}^{\rm ur}_{i}$ and confidence maps $\mathbf{u}^{\rm ur}_{i}$. 
Specifically, starting from random Gaussian noise $\bm{\epsilon}$, we can obtain a clear prediction for a rain video by gradually reversing the diffusion process (referring to Eqs~\eqref{eqn_creverse1} and~\eqref{eqn_creverse2}).
Ideally, different initial Gaussian noise conditions with a rain video input should result in the same clear prediction. However, due to the complexity of rain streak distributions, the predictions from different noise maps might differ in challenging regions, such as those with complex light effects.

Motivated by this, we devise a resampling strategy to derive confidence maps for our rain removal predictions: We randomly generate $N$ noise maps for each training video. Through a gradual reversal of the diffusion process, we obtain $N$ predictions. The final prediction $\mathbf{y}^{\rm ur}_{i}$ is the mean of these predictions, while the confidence map $\mathbf{u}^{\rm ur}_{i}$ signifies the variance among them. These two processes can be defined as follows:%
\begin{eqnarray}
	\label{Eqn_mean}
	\mathbf{y}^{\rm ur}_{i} = \frac{1}{N}\sum^{N}_{n=1}\mathbf{y}^{\rm ur}_{i,n},\\
	\label{Eqn_var}
	\mathbf{u}^{\rm ur}_{i} = {\rm var}\{\mathbf{y}^{ur}_{i,n}\}^{N}_{n=1}.
\end{eqnarray}

Subsequently, we convert confidence maps to a binary mask by setting a threshold $\mathbf{t}_u$. The values $\rm 0$ and $\rm 1$ of the binary mask represent low-confidence and high-confidence regions, respectively. Fig. \ref{fig_vis_self_enhance} demonstrates that the high-confidence regions (blue regions) effectively remove most rain streaks.
However, some high-confidence regions, like the sky, appear dark. This occurs because our confidence maps solely focus on the impact of rain streak removal. Consequently, the confidence scores remain high even when the rain-free background is dark. This issue can be addressed through our adaptive-correction process later.

Finally, we select high-confidence predictions and their corresponding inputs to construct region-based paired real data $\mathbf{D}_{ur} = \{ \mathbf{\widehat{x}}^{ur}_{i}, \mathbf{\widehat{y}}^{ur}_{i} \}^{N_{ur}}_{i=1}$. These real-world pairs are then used to retrain our student model $w_{S}$. 
In each training step, we randomly augment the input videos by adding Gaussian noise and masking pixels. We then feed them into the student model $w_{S}$. 
Training our student model on these augmented inputs  enables it to learn more challenging regions. 
After each optimization of the student model, we update the teacher model via Exponential Moving Average (EMA) that can be formulated as
\begin{equation}
\label{eqn_ema}
w_{T} = 0.999*w_{T} + 0.001*w_{S},
\end{equation}
where $w_{T}$ and $w_{S}$ represent the weights of the teacher and student models, respectively. Through iterative learning, our adaptive-rain-removal enables our model to derain real-world rain videos, particularly in regions affected by complex light effects.

\begin{figure}[t]
	\centering
	{\includegraphics[width=0.32\linewidth]{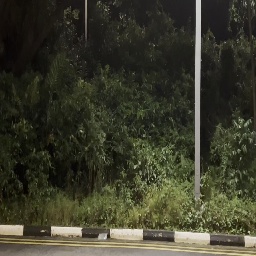}
		\includegraphics[width=0.32\linewidth]{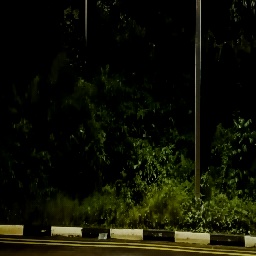}
		\includegraphics[width=0.32\linewidth]{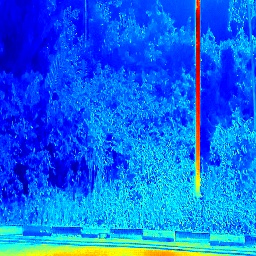}
	}
	{\includegraphics[width=0.32\linewidth]{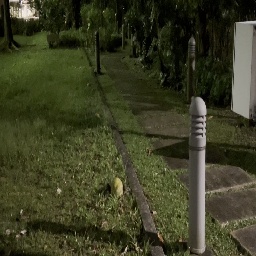}
		\includegraphics[width=0.32\linewidth]{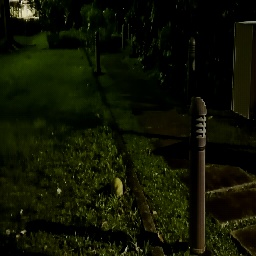}
		\includegraphics[width=0.32\linewidth]{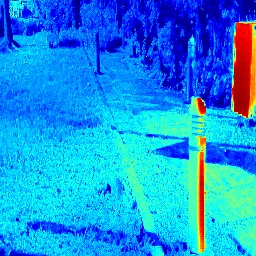}
	}\\
	\hspace{0.3cm} (a) Input  \hspace{0.8cm} (b) Predictions \hspace{0.5cm} (c) Differences
	\\
	\caption{
		Visualization of predictions from our adaptive-rain-removal. (a) Input images. (b) Predictions. (c) Difference maps are calculated using the L1 distance between input images and predictions. 
		The red and blue regions are high-difference and low-difference regions, respectively.
	}
	\label{fig_vis_self_refine}
\end{figure}

\subsection{Adaptive-Correction}
\label{method_srf} 
As our pretrained model relies on a synthetic dataset for training, it tends to produce over-saturated and color-shifted results, as shown in Fig.~\ref{fig_vis_self_enhance}. Our adaptive-correction aims to address these issues, namely restoring the over-saturated and color-shifted regions. The core idea of our adaptive-correction is to rectify errors in our model's predictions by utilizing nighttime clear videos.%

Our adaptive-correction is also implemented based on a teacher-student framework. The parameters of both teacher and student networks are shared in our adaptive-rain-removal and adaptive-correction. 
Given a real-world clear dataset $\mathbf{D}_{uc} = \{ \mathbf{y}^{uc}_{i}\}^{N_{uc}}_{i=1}$, where $\mathbf{y}^{uc}_{i}$ is the $i$-th clear video and $N_{uc}$ is the number of clear videos, we utilize the teacher model $w_{T}$ to generate predictions $\mathbf{x}^{uc}_{i}$. 
Specifically, we first randomly sample a Gaussian noise $\bm{\epsilon}$. Then, we feed the noise and a clear video condition into our teacher model $w_{T}$. By gradually reversing the diffusion process (referring to Eqs~\eqref{eqn_creverse1} and~\eqref{eqn_creverse2}), our teacher model generate restored predictions. In theory, the prediction should be the same as the inputs since the input videos are already clear. 

However, due to the lack of paired real-world data, our initial synthetic-trained diffusion model is difficult to model accurate clear distribution, leading to some over-suppressed regions in our predictions. Fig. \ref{fig_vis_self_refine} shows the input clear night videos and their corresponding predictions. We also visualize the L1 distance between them. The blue regions are low-difference regions, while the red regions represent high-difference regions. It can be observed that predictions obtained using our pretrained video diffusion model are different from their corresponding inputs. These differences essentially represent over-saturated and color-shifted problems and need to be addressed.

Motivated by this, we utilize these differences to construct paired real-world samples $\mathbf{D}_{uc} = \{ \mathbf{\widehat{x}}^{uc}_{i}, \mathbf{\widehat{y}}^{uc}_{i} \}^{N_{uc}}_{i=1}$. We then retrain our student model $w_{S}$ using these differing regions, thereby compelling our student model to rectify these errors. To be specific, the input and ground-truth videos are $\mathbf{\widehat{x}}^{uc}_{i}$ and $\mathbf{\widehat{y}}^{uc}_{i}$, respectively. In each training step, we randomly sample noise $\bm{\epsilon}_t$ with a time step $t$ and add it to the ground-truth $\mathbf{\widehat{y}}^{uc}_{i}$. We subsequently feed the noise-affected ground truth and $\mathbf{\widehat{x}}^{uc}_{i}$ into our noise estimation network for training. The objective function of the noise estimation network is to estimate accurate noise maps $\bm{\epsilon}_t$. This process can be formulated as follows:
\begin{equation}
{\rm Loss} = \vert\vert\bm{\epsilon}_t - \bm{\epsilon}_{w_S}(\sqrt{\bar{\alpha}_t}\mathbf{\widehat{y}}^{uc}_{i}+\sqrt{1-\bar{\alpha}_t}\bm{\epsilon}_t\,\mathbf{\widehat{x}}^{uc}_{i},t)\vert\vert^2,
\end{equation}
where $\alpha_t=1-\beta_t$, $\bar{\alpha}_t=\prod_{i=1}^t\alpha_i$ and $\bm{\epsilon}_{w_S}(\cdot)$ is the noise estimation network of the student model. $\beta_1,\ldots,\beta_T$ are the variance changing schedule.
By continually training on these differences, our student model can rectify errors from our teacher model. After each optimization of the student model, we update our teacher model via EMA (referring to Eq.~(\ref{eqn_ema})). This allows our teacher model to avoid producing erroneous results, leading to more precise restored results.

In the training stage, we jointly use adaptive-rain-removal and adaptive-correction to train our student model $w_{S}$. The two processes share the same teacher and student parameters. After each optimization of the student model, we use EMA to update the teacher model $w_{T}$. In the inference stage, given a degraded video, we directly use the teacher model $w_{T}$ to generate rain-free predictions. Note that, $w_{T}^{\rm SR}$ = $w_{T}^{\rm SC}$ and $w_{S}^{\rm SR}$ = $w_{S}^{\rm SC}$, where $T$ represents the teacher model, $S$ represents the student model, $\rm SR$ stands for adaptive-rain-removal and $\rm SC$ stands for adaptive-correction. The network parameters are always shared.

\begin{table*}[t]
	\centering
	\renewcommand{\arraystretch}{1.2}
		\resizebox{0.99\textwidth}{!}{
			\begin{tabular}{c|c|c|c|c|c|c|c|c}
				\toprule
				\multicolumn{1}{c|}{Methods} & \multicolumn{1}{c|}{HRIR} & \multicolumn{1}{c|}{DLF} & \multicolumn{1}{c|}{RMFD} & \multicolumn{1}{c|}{DSTFM} & \multicolumn{1}{c|}{MetaRain} & \multicolumn{1}{c|}{WeatherDiff} & \multicolumn{1}{c|}{FDM} & \multirow{2}[2]{*}{Ours} \\
				\multicolumn{1}{c|}{Venue} & \multicolumn{1}{c|}{CVPR'19} & \multicolumn{1}{c|}{CVPR'19} & \multicolumn{1}{c|}{TPAMI'21} & \multicolumn{1}{c|}{PR'22} & \multicolumn{1}{c|}{ECCV'22} & \multicolumn{1}{c|}{TPAMI'23} & \multicolumn{1}{c|}{NeurIPS'22} &  \\
				\midrule
				PSNR  & 16.83 & 15.71 & 16.18 & 17.82 & 22.21  & 20.98   & 23.49  & \textbf{26.73} \\
				\hline
				SSIM  & 0.6481 & 0.6307 & 0.6402 & 0.6486 & 0.6723  & 0.6697   & 0.7657  & \textbf{0.8647} \\
				\bottomrule
			\end{tabular}%
		}
	\caption{Quantitative results on the nighttime deraining dataset. DSTFM~\cite{patil2022dual} and MetaRain~\cite{patil2022video} are nighttime video deraining methods. WeatherDiff~\cite{ozdenizci2023restoring} and FDM~\cite{harvey2022flexible} are diffusion-based methods, where FDM is a video diffusion model.
	}
		\label{tab_syn}%
	\end{table*}%

\begin{figure*}[t]

\subfloat[Input Frames]
{\includegraphics[width=2.8cm, height=2.8cm]{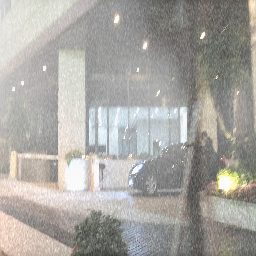}}\hspace{1pt}
\subfloat[Ground-Truth]
{\includegraphics[width=2.8cm, height=2.8cm]{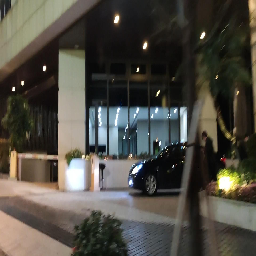}}\hspace{1pt}
\subfloat[\textbf{Ours}]
{\includegraphics[width=2.8cm, height=2.8cm]{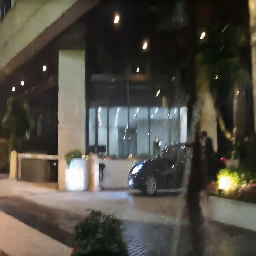}}\hspace{1pt}
\subfloat[MetaRain]
{\includegraphics[width=2.8cm, height=2.8cm]{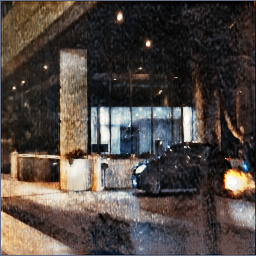}}\hspace{1pt}
\subfloat[WeatherDiff]
{\includegraphics[width=2.8cm, height=2.8cm]{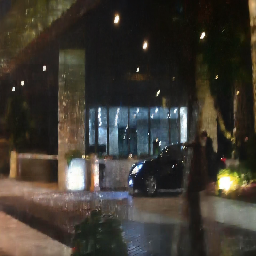}}\hspace{1pt}
\subfloat[FDM]
{\includegraphics[width=2.8cm, height=2.8cm]{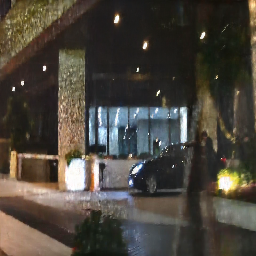}}\hspace{1pt}

\caption{Qualitative results on the SynNightRain dataset. MetaRain \cite{patil2022video} is a nighttime video deraining method. WeatherDiff~\cite{ozdenizci2023restoring} and FDM~\cite{harvey2022flexible} are diffusion-based methods, where FDM is a video diffusion method. Zoom in for better visualisation. }
\label{fig_qual_syn}
\end{figure*}

	\section{Experiments}
	
	\noindent \textbf{SynNightRain} \cite{patil2022video, patil2022dual} is a synthetic nighttime video deraining dataset. 	
	It includes 30 nighttime videos, each of which includes 200 frames.
	Each video contains synthetic rain streaks that generated by \cite{ranftl2020towards} and \cite{li2019heavy}.
	We follow the  protocol~\cite{patil2022video} to evaluate the effectiveness of our method, i.e., 10 videos are used as the training set and the rest 20 videos are taken as the test set. 
	
	\noindent \textbf{RealNightRain} To evaluate the generalization ability of our method,
	we collect 10 real-world nighttime rain videos, where 5 videos contain dense rain streaks and the rest 5 videos contain sparse rain streaks. 
	We will publicly release our collected dataset.
	
	\subsection{Implementation Details}
	Our NightRain includes three parts: pretraining, adaptive-rain-removal and adaptive-correction, each of which needs to train a video diffusion model.
	In each training step, we randomly sample $\rm P \times K$ videos, where $\rm P$ is the number of videos and $\rm K$ denotes the number of clips of a video. 
	Each clip consists of 4 frames with the image size is $64\times 64$. 
	The Adam is used to optimize our model and the learning rate is set to 0.0002. 
	%
	%
	%
	%
	
	\noindent \textbf{Pretraining} 
	The total training steps and $\rm P\times K$ are set to 2,000,000 and $16\times 4$, respectively. 
	To evaluate our method on SynNightRain, we use 25 sampling steps to gradually restore a rain video. 
	
	\noindent \textbf{Adaptive-Rain-Removal} Fig. \ref{Fig_overview_srr} shows the pipeline of our adaptive-rain-removal. 
	The number of iterations in confidence generation $N$, the threshold $\rm t_{u}$ and the total training steps are set to 3, 0.5 and 10,000, respectively. 
    We apply two data augmentation approaches (Gaussian noise and pixel masking) to augment the inputs of the student model. The variance range of the Gaussian noise is $(0, 0.2)$ and the masking ratio of pixels is set to 25\%.
    The total training steps and $\rm P\times K$ are set to 10,000 and $10\times 2$, respectively. 
	
	\noindent \textbf{Adaptive-Correction} Fig. \ref{Fig_overview_sc} shows the pipeline of our adaptive-rain-removal. The total training steps, the number of clear videos, $\rm P$ and $\rm K$ are set to 10K, 10, 10 and 2, respectively.
	
	\noindent \textbf{Transformer-based Video Diffusion Model} 
	As shown in Figure \ref{Fig_transformer}, our video diffusion model consists of three modules, i.e., Patch Partition, Time Embedding and Transformer blocks. 
	For the patch partition part, we set the patch size $(\rm ts\times ss\times ss)$ and the output channel $C_{p}$ to 2 and 768, respectively. 
	For the time embedding, we utilize two fully-connected layers to encode the information, where the output channels are set to 768.
    In our method, we set the number of transformer blocks to 10. 
    %

\begin{figure*}[t]
{\includegraphics[width=3.4cm, height=3.4cm]{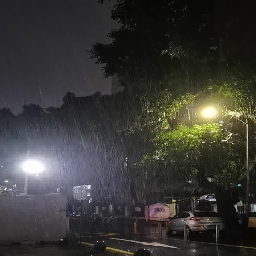}}\hspace{1pt}
{\includegraphics[width=3.4cm, height=3.4cm]{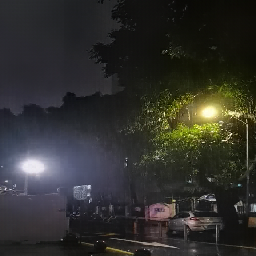}}\hspace{1pt}
{\includegraphics[width=3.4cm, height=3.4cm]{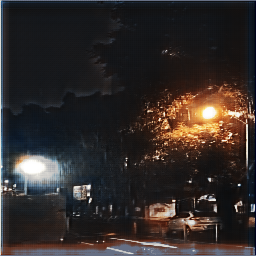}}\hspace{1pt}
{\includegraphics[width=3.4cm, height=3.4cm]{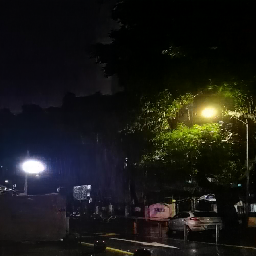}}\hspace{1pt}
{\includegraphics[width=3.4cm, height=3.4cm]{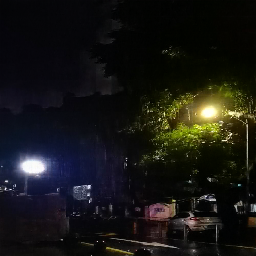}}\hspace{1pt}

\subfloat[Input Frames]
{\includegraphics[width=3.4cm, height=3.4cm]{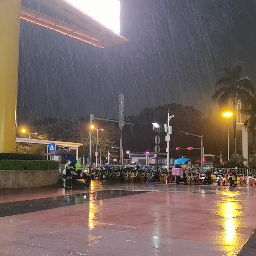}}\hspace{1pt}
\subfloat[\textbf{Ours}]
{\includegraphics[width=3.4cm, height=3.4cm]{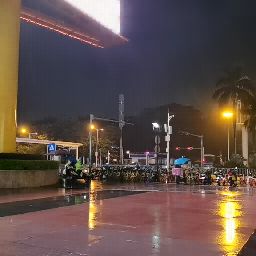}}\hspace{1pt}
\subfloat[MetaRain]
{\includegraphics[width=3.4cm, height=3.4cm]{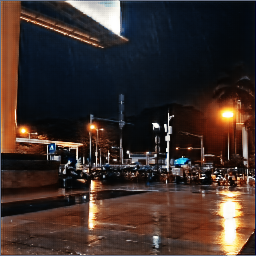}}\hspace{1pt}
\subfloat[FDM]
{\includegraphics[width=3.4cm, height=3.4cm]{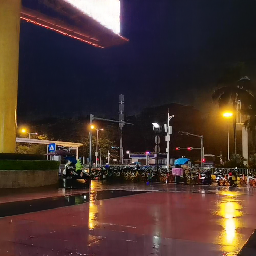}}\hspace{1pt}
\subfloat[WeatherDiff]
{\includegraphics[width=3.4cm, height=3.4cm]{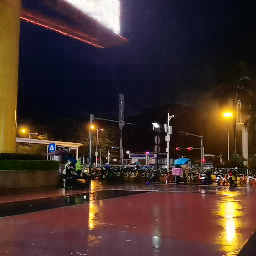}}\hspace{1pt}

\caption{Qualitative comparison on real-world data. MetaRain \cite{patil2022video} is nighttime video deraining methods. FDM~\cite{harvey2022flexible} and WeatherDiff~\cite{ozdenizci2023restoring} are diffusion-based methods, where FDM is a video diffusion method. Zoom in for better visualisation. }
\label{fig_qual_real1}
\end{figure*}

 \begin{figure}[t]
{\includegraphics[width=2cm, height=2cm]{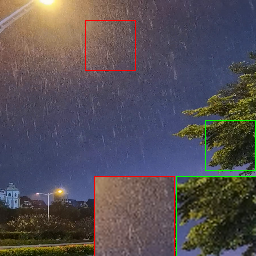}}\hspace{1pt}
{\includegraphics[width=2cm, height=2cm]{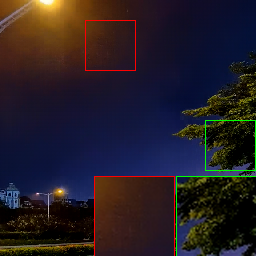}}\hspace{1pt}
{\includegraphics[width=2cm, height=2cm]{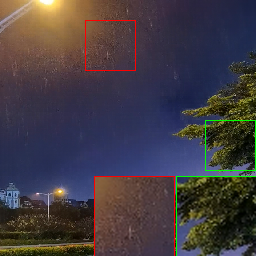}}\hspace{1pt}
{\includegraphics[width=2cm, height=2cm]{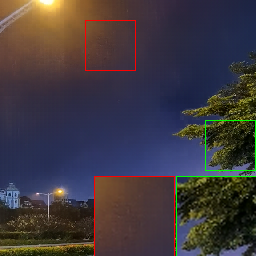}}\hspace{1pt}

\subfloat[Inputs]
{\includegraphics[width=2cm, height=2cm]{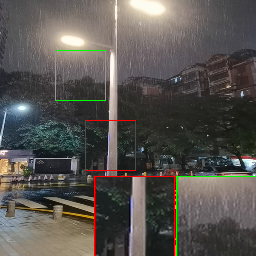}}\hspace{1pt}
\subfloat[Pre-trained]
{\includegraphics[width=2cm, height=2cm]{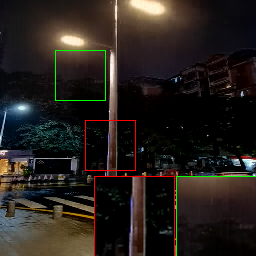}}\hspace{1pt}
\subfloat[AC]
{\includegraphics[width=2cm, height=2cm]{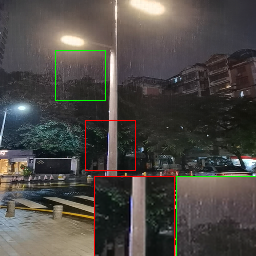}}\hspace{1pt}
\subfloat[AC + AR]
{\includegraphics[width=2cm, height=2cm]{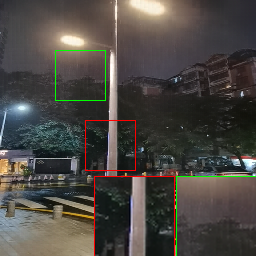}}\hspace{1pt}

\caption{Ablation studies on real-world videos. (a) Input  (b) Results obtained using the pre-trained model. (c) Results obtained using our adaptive-correction. (d) Results obtained using our adaptive-correction and adaptive-rain-removal. Zoom in for better visualisation.
}
\label{fig_qual_Ablation}
\end{figure}

	\subsection{Evaluation on the SynNightRain Dataset}
	We compare our NightRain on the SynNightRain dataset with state-of-the-art (SOTA) methods, including HRIR~\cite{li2019heavy}, DLF~\cite{yang2019frame}, RMFD~\cite{yang2021recurrent}, DSTFM~\cite{patil2022dual}, MetaRain~\cite{patil2022video}, FDM~\cite{harvey2022flexible} and WeatherDiff~\cite{ozdenizci2023restoring}. HRIR, DLF, RMFD, DSTFM and MetaRain are CNN-based methods, while the rest methods are diffusion-based methods. The quantitative results of HRIR, DLF, RMFD and DSTFM are collected from~\cite{yang2021recurrent}.
	
	\noindent \textbf{Quantitative Evaluation} The experimental results are shown in Table \ref{tab_syn}. 
	It can be observed that the PSNR and SSIM of MetaRain are 22.21 and 0.6723, respectively. 
	Our method achieves 26.73 of PSNR and 0.8647 of SSIM, which outperforms MetaRain by 4.52 dB and 0.1924, respectively.
	We also compare our method with existing state-of-the-art diffusion methods, i.e., FDM~\cite{harvey2022flexible} and WeatherDiff~\cite{ozdenizci2023restoring}.
	All experiments use the same sampling steps for a fair comparison. 
	Our NightRain outperforms these methods by 3.24 of PSNR.
	This is because existing methods are implemented by convolution neural networks and thus cannot take full advantage of global spatiotemporal information.
	In contrast, we develop a transformer-based diffusion model that fully utilizes global spatiotemporal cues. 
	As a result, our NightRain can estimate more accurate noise maps, leading to superior performance.
	
	\noindent \textbf{Qualitative Evaluation} Fig. \ref{fig_qual_syn} shows the quantitative results, including MetaRain~\cite{patil2022video}, FDM~\cite{harvey2022flexible}, WeatherDiff~\cite{ozdenizci2023restoring} and our NightRain. 
	We can observe that existing methods struggle to restore video details in dense rain streaks since these methods do not effectively model clear distributions and capture global information.
	In contrast, our NightRain introduces a transformer-based diffusion model that captures global relations and models more comprehensive clear distributions. As a result, it effectively recovers details from adjacent pixels, leading to superior performance.

	\subsection{Evaluation on Real-world Datasets} Fig. \ref{fig_qual_real1} shows the qualitative results on real-world nighttime rain videos. 
	It can be observed that the restored results of existing nighttime video deraining methods still have a few rain streaks left. 
	This is because these methods are trained on synthetic datasets and thus suffer from the domain gap problem. 
	In contrast, our NightRain not only removes nighttime rain streaks but also restores the clear background.
	The main reason is that our adaptive-rain-removal utilizes high-confidence predictions from real-world rain videos to improve the deraining ability of our model. By training on actual paired data, our model reduces the domain gap problem between synthetic and real-world rain streak distributions. 

	Another problem of existing methods is that their restored predictions are over-saturated or suffer from a color shift (refer Figure \ref{fig_trailer}). For example, MetaRain~\cite{patil2022video} over-suppresses the white light. The results of FDM~\cite{harvey2022flexible} and WeatherDiff~\cite{ozdenizci2023restoring} are too dark. This problem is caused by the domain gap between synthetic and real-world rain streak distributions. 
    Our NightRain also relies on synthetic datasets to initialize our video diffusion model. However, we propose adaptive-correction to address these problems. Specifically, adaptive-correction forces our model to continually rectifies prediction errors during the training stage. As a result, our NightRain effectively removes rain streaks without introducing over-saturation or color-shift issues, resulting in more accurate detail recovery.

	\subsection{Ablation Studies}

    In this section, we conduct several ablation studies on real-world datasets to verify the effectiveness of our proposed processes: adaptive-rain-removal and adaptive-correction. Figure \ref{fig_qual_Ablation} displays the results of the pre-trained model, our adaptive-rain-removal, and adaptive-correction. Moreover, we use Hyper-IQA \cite{su2020blindly}, a non-reference metric, to quantitatively evaluate the performance. The Hyper-IQA scores (higher is better) for (a) Inputs, (b) Pre-trained, (c) AC, and (d) AC + AR are 0.6341, 0.6401, 0.6540, and 0.6791, respectively. 
    It can be observed that predictions obtained using the pre-trained model suffer from over-saturation and color shifts. Specifically, the results are excessively dark, hindering the visibility of object details. This is due to the significant domain gap between synthetic and real-world rain streak distributions.

    As illustrated in Figure \ref{fig_qual_Ablation} (c), our adaptive-correction addresses over-saturation and color-shift issues by continually rectifying errors in our predictions. However, despite these improvements, some rain streaks remain in the predictions of our adaptive-correction. To tackle this, our adaptive-rain-removal utilizes unlabeled rain videos, enabling our model to remove rain streaks from real-world rain videos, especially in regions affected by complex light effects. Consequently, the final outputs of our NightRain not only eliminate rain streaks but also retain crucial video details.

\section{Conclusion}
We have presented NightRain, a novel nighttime video deraining method with adaptive-rain-removal and adaptive-correction processes.
Our adaptive-rain-removal utilizes unlabeled rain videos to enable our model to enhance the deraining capability in real-world rain videos, particularly in regions affected by complex light effects. By learning high-confidence predictions, this process compels our model to progressively address more challenging regions, leading to effective real-world rain streak removal.
Our adaptive-correction aims to address over-saturation and color shift problems. By learning from differences between clear night videos and their corresponding predictions, our adaptive-correction continually rectifies errors in our models’ predictions, resulting in more precise restoration.
Extensive experiments show that our NightRain achieves a significant performance improvement both quantitatively and qualitatively.


\bibliography{aaai24}

\end{document}